\documentclass{article}
\pdfoutput=1

    \PassOptionsToPackage{numbers, compress}{natbib}
\usepackage[preprint]{neurips_2023}

\usepackage[utf8]{inputenc} % allow utf-8 input
\usepackage[T1]{fontenc}    % use 8-bit T1 fonts
\usepackage{url}            % simple URL typesetting
\usepackage{booktabs}       % professional-quality tables
\usepackage{amsfonts}       % blackboard math symbols
\usepackage{nicefrac}       % compact symbols for 1/2, etc.
\usepackage{microtype}      % microtypography
\usepackage{xcolor}         % colors
\usepackage{epsfig}
\usepackage{graphicx}
\usepackage{amsmath}
\usepackage{amssymb}
\usepackage[pagebackref,breaklinks,colorlinks]{hyperref}
\newcommand{\ie}{\textit{i.e.}}
\newcommand{\eg}{\textit{e.g.}}

\title{Caption Anything:  Interactive Image Description \\ with Diverse Multimodal Controls}

\author{Teng Wang$^{*\dagger}$, Jinrui Zhang$^{*}$, Junjie Fei$^{*}$, Hao Zheng, Yunlong Tang, Zhe Li,\\ \textbf{Mingqi Gao, Shanshan Zhao}\\
SUSTech VIP Lab
}

\begin{document}
\maketitle

%%%%%%%%% ABSTRACT
\begin{figure*}[hb]
\centering
\vspace{-2em}
    \includegraphics[width=0.98\textwidth]{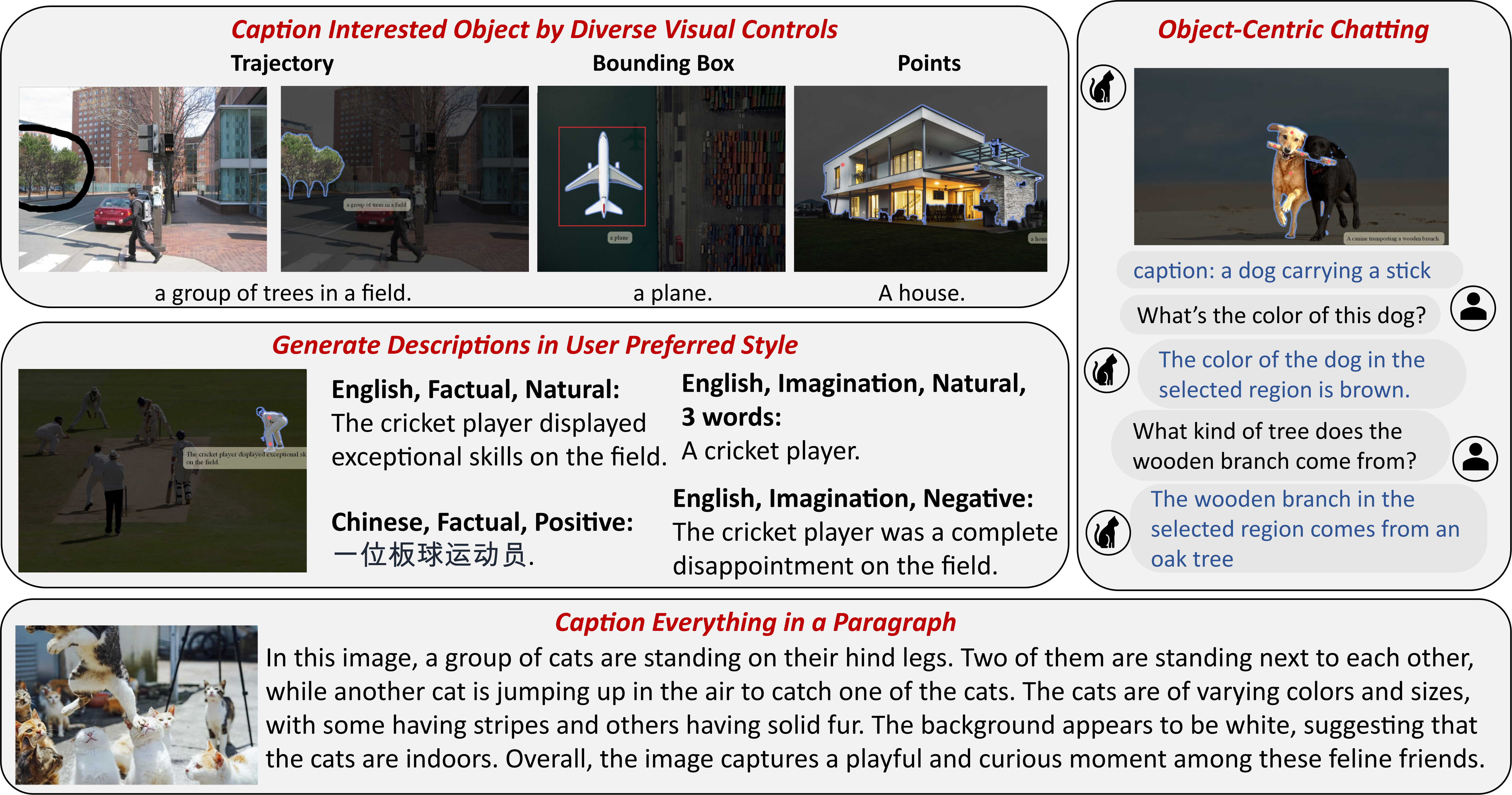}
    \vspace{-0.5em}
    \caption{Caption Anything supports a diverse range of visual and language controls, making it effortlessly adaptable for object-centric chatting and image paragraph captioning.}
    % \vspace{-1.0 em}
    \label{fig:app}
\end{figure*}

\begin{abstract}

\let\thefootnote\relax\footnotetext{$^*$ Equal contribution. $^{\dagger}$ Work done during  internship in ARC Lab, Tencent PCG. We thank Yixiao Ge and Ying Shan for their support and constructive discussions.}

Controllable image captioning is an emerging multimodal topic that aims to describe the image with natural language following human purpose, \eg, looking at the specified regions or telling in a particular text style. State-of-the-art methods are trained on annotated pairs of input controls and output captions. However, the scarcity of such well-annotated multimodal data largely limits their usability and scalability for interactive AI systems. Leveraging unimodal instruction-following foundation models is a promising alternative that benefits from broader sources of data. In this paper, we present Caption AnyThing (CAT), a foundation model augmented image captioning framework supporting a wide range of multimodel controls: 1) visual controls, including points, boxes, and trajectories; 2) language controls, such as sentiment, length, language, and factuality. Powered by Segment Anything Model (SAM) and ChatGPT, we unify the visual and language prompts into a modularized framework, enabling the flexible combination between different controls. Extensive case studies demonstrate the user intention alignment capabilities of our framework, shedding light on effective user interaction modeling in vision-language applications. Our code is publicly available at \url{https://github.com/ttengwang/Caption-Anything}.

\end{abstract}

\section{Introduction}
\begin{figure*}
\centering
    \includegraphics[width=1.0\textwidth]{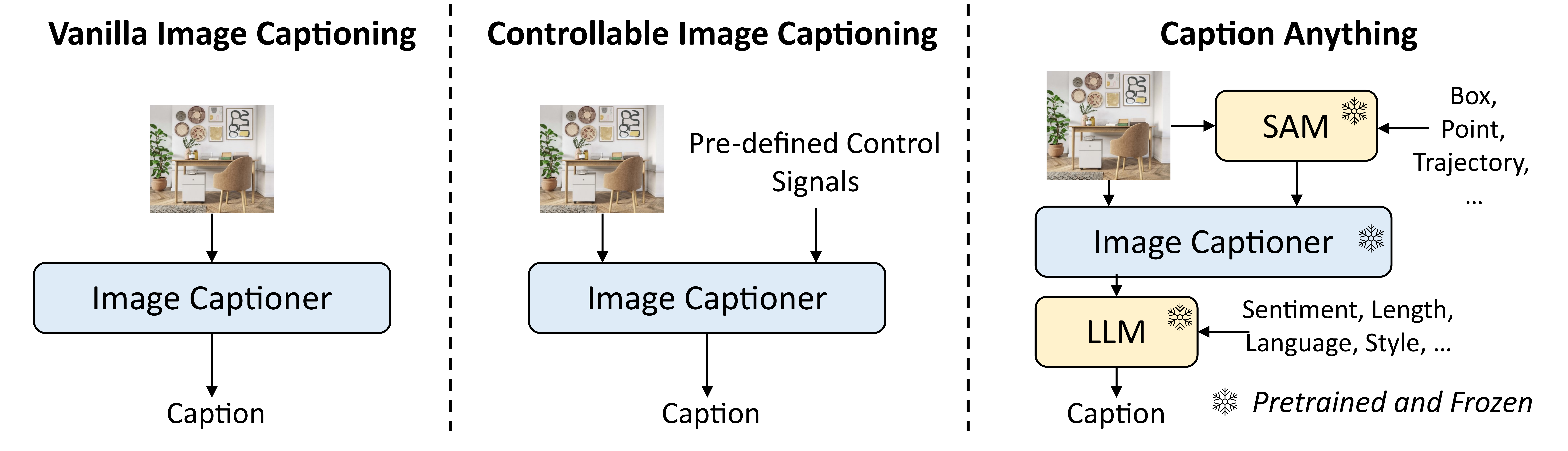}
    % \vspace{-1.0emm}
    \caption{Comparison between image captioning pipelines. Vanilla image captioning methods lack explicit control, making them unsuitable to interact with users. Previous controllable image captioning methods mainly rely on limited-scale human-annotated data with specific control signals, and they only support pre-defined control signals. The proposed CAT is training-free and supports diverse visual controls and language controls.}
    \vspace{-1.0 em}
    \label{fig:intro}
\end{figure*}

Describing images in natural language is a critical problem of vision-language learning. Fig.~\ref{fig:intro} demonstrates the difference between image captioning systems. Vanilla image captioning~\cite{hu2022scaling, fang2022injecting, li2022comprehending, fei2022deecap, wang2023accelerating} and dense captioning~\cite{johnson2016densecap, yang2017dense, wu2022grit} typically present an image's salient features using objective and pragmatic language, either in a single sentence or a set of sentences. However, such approaches may produce output that is excessively concise or overly complex, rendering them unsuitable for user interaction due to the lack of explicit control that aligns with user intention. Controllable Image Captioning (CIC) is a promising research direction that aligns language output with user intent. Subsequent methods~\cite{deng2020length, chen2021human, cornia2019show, pont2020connecting} have been proposed to incorporate diverse control signals into image captioning models. However, their applicability is limited by two primary factors: (1) Existing CIC models typically rely on training with human-annotated (image, text, control signal) tuples~\cite{gan2017stylenet, mathews2016senticap}. The limited scale of the datasets constrains these models' capacity to comprehend control signals; (2) These models only support pre-defined single or several control signals, which limits their flexibility of combining different controls and introducing new dimensions of controllability.

To tackle these issues, we propose Caption AnyThing (CAT), a zero-shot controllable image captioning framework augmented by pre-trained foundation models. Specifically, CAT integrates pre-trained image captioners~\cite{li2022blip, li2023blip2, wang2022git} with SAM~\cite{kirillov2023segment} and an instruction-tuned LLM. The image and the visual controls are first processed by SAM, which generates a pixel-level mask that corresponds to the selected region, thereby facilitating the perception centered on user-interested objects. Benefiting from the various visual prompts (\eg, points, bounding boxes) used during the training of SAM, CAT supports flexible visual controls when interacting with users. The output sentences are further refined by an instruction-tuned LLM. Prominent LLMs, such as GPT-4~\cite{openai2023gpt4} and FLAN-T5~\cite{chung2022scaling}, are tuned with human feedback, thereby enabling CAT to accommodate a variety of language controls and align more effectively with user intent.

In contrast to existing controllable captioning approaches, CAT leverages foundation models to establish its controllability rather than solely relying on learning pre-defined control signals from training data. This approach not only reduces the reliance on human-annotated data, leading to a training-free model but also enhances the model's transferability by harnessing the knowledge embedded in vast pre-training data. Furthermore, CAT supports a diverse range of control signals, making it highly adaptable and extendable for interactive use. Currently, it supports 3 vision (click, boxes, trajectory) and 4 language (sentiment, length, language, factuality) control signals, which can be flexibly combined to generate diverse and personalized captions. In addition, our model provides unified representations for both two types of controls. To be specific, the visual controls and language controls are unified to pixel-level masks and textual prompts, respectively. With this design, CAT could be readily expanded with any control signals that can be transposed into these unified representations, thereby augmenting its flexibility and scalability.

We present the strong user-interactive capabilities of CAT through a comprehensive array of qualitative examples. As shown in Fig.~\ref{fig:app}, users could select the interested objects via various visual controls to generate captions in their preferred styles. Moreover, by incorporating additional OCR and VQA tools, CAT could be easily extended to two multimodal applications, namely, object-centric chatting and image paragraph captioning. The former enables users to chat around specific objects, facilitating a more in-depth understanding of interested objects. The latter effectively integrates knowledge from various domains originating from distinct foundation models, enabling the generation of detailed and logically coherent descriptions. Overall, diverse cases show that CAT is a highly interactive multimodal system, exhibiting considerable potential for real-world applications.

In summary, the contributions of this paper are three-fold: 1) We propose a training-free CIC framework that is built upon foundation models, leading to reduced reliance on human-annotated data. 2) Our approach supports a more diverse range of controls and offers unified representations for both visual and language controls, facilitating extensibility to incorporate new aspects of controllability 3) Experiments demonstrate strong user-interactive capabilities of CAT.

\section{Related Work}

\paragraph{Image Captioning and Dense Captioning.} Image captioning is a multimodal task that aims to generate descriptions for a given image. The prevailing approaches mainly employ the encoder-decoder paradigm~\cite{vinyals2015show,xu2015show,anderson2018bottom,huang2019attention,li2022blip,li2023blip2, wang2022git} to solve this task. 
% Most of them use carefully designed attention mechanisms to improve the scene understanding ability of the models.
To enrich the detailed understanding for complex scenes, dense captioning~\cite{johnson2016densecap, yang2017dense, li2019learning,yin2019context,shao2022region,wu2022grit} is proposed to generate localized captions for all salient objects in an image. Yin et al.~\cite{yin2019context} propose a multi-scale contextual information-sharing technique to capture fine-grained global context information. Wu et al.~\cite{wu2022grit} propose a unified framework for dense captioning and object detection, achieving better object understanding capability. Nonetheless, these methods generate descriptions based on the image's salient features only, rendering them unsuitable for user interaction as they lack an explicit control mechanism that aligns with the user's intention. Compare with them, our method focuses on interactive image description and supports diverse multimodal controls to provide better user alignment.
% \vspace{-1.0 em}

\paragraph{Controllable Image Captioning.} The controllable capability to generate desired descriptions related to user-specified objects in the image is useful in practical applications. On the one hand, the visual controls in controllable image captioning usually involve the bounding box~\cite{cornia2019show}, mouse behavior~\cite{pont2020connecting,yan2021control}. Cornia et al.~\cite{cornia2019show} propose to generate the corresponding caption based on a sequence or a set of image regions. Pont-Tuset et al.~\cite{pont2020connecting} release an interactive image captioning dataset where the annotators are required to describe the image region the mouse trajectory covered. LoopCAG~\cite{yan2021control} further improves the generation quality of captions and interactive controllability. On the other hand, the text style of the generated caption can be flexibly changed according to different application scenarios. Typically, the controllable text style can be summarized as: length controllability~\cite{deng2020length}, sentiment controllability~\cite{mathews2016senticap}, imaginary controllability (\eg, romantic or humorous descriptions)~\cite{zhao2020memcap,gan2017stylenet}. Some works~\cite{zeng2023conzic,wang2022controllable} unify these controllable styles into a single architecture. BLIP2~\cite{li2023blip2}, a language-image pre-training model, performs zero-shot image-to-text generation following natural language prompts. In this paper, we further extend the flexibility of controllable image captioning, where the interaction can be points, boxes, or trajectory specified by users, and the language style of generated descriptions is able to meet users' requirements to the greatest extent.
% \vspace{-1.0 em}
\paragraph{Interactive Image Segmentation.} Interactive image segmentation~\cite{liew2017regional,wang2018deepigeos,xu2016deep,sofiiuk2020f} is an important research problem where the model is required to segment the image according to the user's intention with interaction (\eg, point, trajectory). Xu et al.~\cite{xu2016deep} propose deep-learning-based interactive image segmentation firstly. Li et al present an end-to-end interactive image segmentation~\cite{li2018interactive}. The whole architecture is divided into two convolutional networks, where the first is used to synthesize various masks according to the user's input, and the second is trained to select a single solution among these masks. FCA-Net~\cite{lin2020interactive} makes better use of the first click to improve the interactive segmentation result. SAM~\cite{kirillov2023segment} builds a promptable foundation model for segmentation, where the prompt can be any information indicating what to segment in an image, \eg, a set of points, a rough box or mask, or free-form text.
% \vspace{-1.0 em}
\paragraph{Large Language Models.}
LLMs have arisen significant interest due to their strong transfer capabilities across a diverse range of language processing tasks, as well as their emerging capabilities to interact with humans. Recently, the most important breakthrough was made by GPT-3~\cite{brown2020language}, a model with 175B parameters, which unveiled the emerging potential of the few-shot learning techniques. This remarkable performance motivated lots of subsequent LLMs~\cite{zhang2022opt, chung2022scaling, touvron2023llama, scao2022bloom}. In order to enhance the interactive capacity of LLMs with human users, state-of-the-art models are commonly fine-tuned with human feedback~\cite{ouyang2022training, alpaca, muennighoff2022crosslingual, openai2023gpt4, taylor2022galactica}. This approach enables LLMs to effectively adapt to diverse language instructions, thereby aligning more closely with user intent. Inspired by the user alignment capabilities of LLMs, we unify language controls to textual prompts and apply LLMs to generate image descriptions that align with user preferences.

\section{Caption Anything}

\begin{figure*}[h]
    \centering
    \includegraphics[width=1.0\textwidth]{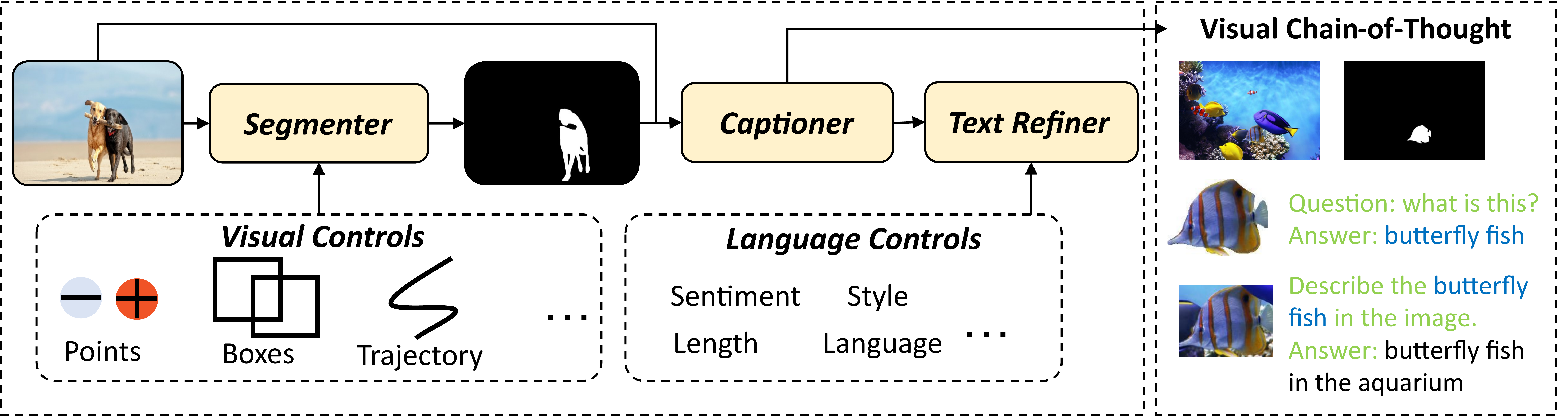}
    \caption{The overall framework of Caption Anything. It introduces multimodal controls to image captioning, rendering a variety of visual focuses and language styles aligned with human intention. The visual prompt is firstly converted into the mask prompt by the \textit{segmenter}.  Subsequently, the \textit{captioner} predicts a raw caption for the region delineated by the mask. To make the \textit{captioner} focus on the user-interested object, we use a simple visual chain-of-thought technique to conduct step-by-step inference. Finally, both the text prompt and the raw caption are fed into the \textit{text refiner}, which generates a user-preferred caption in accordance with the desired genre.}
    \label{fig:demo}
\end{figure*}

To enhance the user-centric interactivity of current image captioning systems, we propose a foundational model augmentation strategy to accommodate  image captioners with a variety of multimodal controls. Specifically, our approach could be formulated as a triplet solver~\{\textit{segmenter}, \textit{captioner}, \textit{text refiner}\}. As illustrated in Fig.~\ref{fig:demo}, \textit{segmenter} 
 first takes the interactive visual controls (\eg, points, boxes, trajectory) and represents the user-interested regions via pixel-level masks. Subsequently, the \textit{captioner} generates raw descriptions in relation to the specified region based on the original image and the provided mask. In order to facilitate the \textit{captioner} focus on the user-interested object, we design a visual chain-of-thought technique with step-by-step inference. Lastly, \textit{text refiner} refines the raw descriptions by incorporating user-defined language controls, thereby tailoring the language style according to user preferences.

% \vspace{-1.0 em}
\paragraph{Segmenter.}
The ideal \textit{segmenter} is capable of segmenting any part of an image according to the visual controls. SAM~\cite{kirillov2023segment} meets the requirements well and has impressive zero-shot transferability to new image domain, benefiting from promptable pre-training and the SA-1B dataset~\footnote{The project link: \url{https://segment-anything.com}} (the largest segmentation dataset with 1 billion masks on 11M images). SAM adapts interactive segmentation~\cite{mahadevan2018iteratively} to achieve promptable ability, where a prompt, any interaction (\eg, points, boxes) indicating what to segment in an image, is used to prompt SAM to return a valid segmentation mask. Once we obtain the user-specified segmentation mask, it is easy to generate the desired caption according to the original image and mask prompt.

% \vspace{-1.0 em}
\paragraph{Captioner.}
To describe any user-specific object in the image, the \textit{captioner} is expected to perform strong zero-shot captioning performance. In other words, the ideal \textit{captioner} should generate reasonable descriptions among various novel objects and different image distributions. 
We use BLIP2 as the captioner. It leverages frozen pre-trained image encoders and frozen LLMs together with a querying transformer to bridge the modality gap, achieving excellent zero-shot performance.
% Two alternative \textit{captioner} (\ie, BLIP~\cite{li2022blip}, BLIP2~\cite{li2023blip2}, GIT~\cite{wang2022git}) in this paper. GIT unifies vision-language tasks with only one image encoder and one text decoder. BLIP proposes CapFilt to achieve performance improvement by bootstrapping the captions. 
% \vspace{-1.0 em}

\paragraph{Text Refiner.}
In most cases, image-related descriptions should follow users' preferences. However, refining the raw caption generated by the \textit{captioner} based on user instructions is a nontrivial task. To achieve this goal, we introduce ChatGPT as an API to generate more expressive and controllable descriptions from the raw caption. The \textit{text refiner} can be replaced with open-source alternative LLMs easily, such as LLaMA~\cite{touvron2023llama}, OPT-IML~\cite{zhang2022opt}, BLOOM~\cite{scao2022bloom}.

% \vspace{-1.0 em}
\paragraph{Visual Chain-of-Thought.}
We empirically found that the generated object captions are easily affected by the background information. Inspired by the chain-of-thought (CoT) prompting originated from NLP~\cite{wei2022chain,kojima2022large}, we bootstrap the \textit{captioner} with step-by-step text generation to ensure that the generated description focuses on the user-selected region. As Fig.~\ref{fig:demo} shows, we first retain the user-selected object and replace the background with white and ask the \textit{captioner} to identify the category of the interested object. Subsequently, the \textit{captioner} takes the generated text and the cropped image with the background as a prompt to generate the final caption. In this way, the \textit{captioner} is capable of focusing on the selected object during the caption generation process.

% \vspace{-1.0 em}
\paragraph{Extension to Object-Centric Chatting.}
Drawing inspiration from recent advancements in multimodal dialog systems, we investigate the potential of visual dialog specifically targeting objects identified by visual prompts. Unlike dominant chat systems that are designed for a global comprehension of the entire image, local region chat has broader applications for complex, informative, and high-resolution images. This includes visual navigation, embodied intelligence, and early-child education. Given the segmentation mask of an object and user query, we use an off-the-shelf visual question answering model~\cite{li2023blip2} as a visual API that empowers the ChatGPT to understand detailed visual cues by asking questions. Specifically, we include the generated caption to the initial prompt and follow~\cite{wu2023visual} to use LangChain~\cite{Chase_LangChain_2022} as the control hub to predict the chain of API calls.

\paragraph{Extension to Paragraph Captioning.} Our framework is adaptable to the image paragraph captioning task by leveraging ChatGPT to summarize the dense captions and scene texts of an image into a paragraph. Specifically, we generate dense captions by initially using the SAM to segment everything within the image, followed by captioning each object with the CAT pipeline. To incorporate scene text information into the paragraph, we utilize additional OCR tools (\eg, EasyOCR~\cite{EasyOCR}) to identify the text present in the image. The dense captions and scene texts are subsequently merged into a predefined prompt template, which is then employed to instruct ChatGPT in summarizing the scene information into a cohesive paragraph. 
\section{Experiments}
\begin{figure*}
    \centering
    \includegraphics[width=1.0\textwidth]{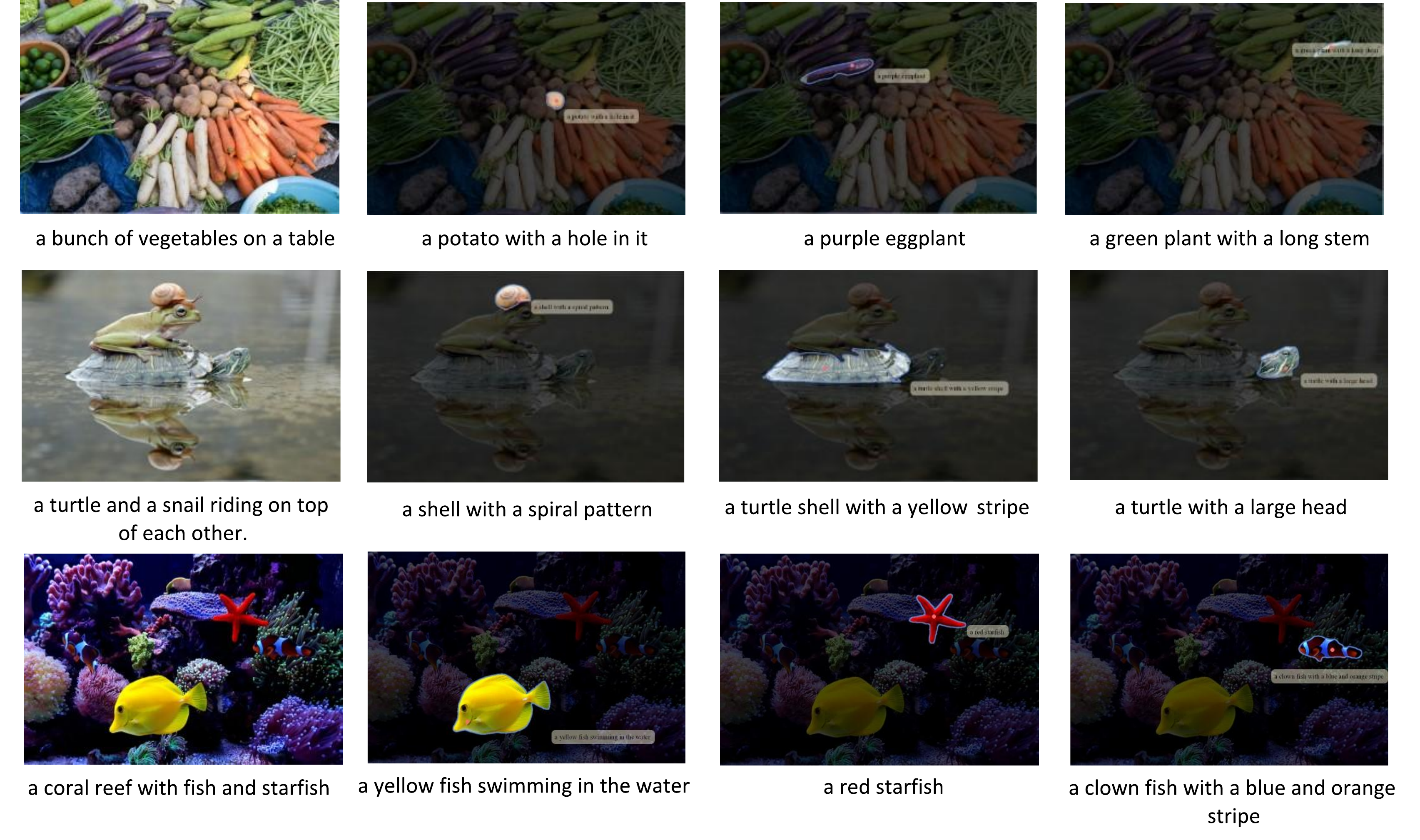}
    \caption{Visualization of describing the image with point-based visual controls.}
    \label{fig:example1}
\end{figure*}
\begin{figure*}
    \centering
    \includegraphics[width=1.0\textwidth]{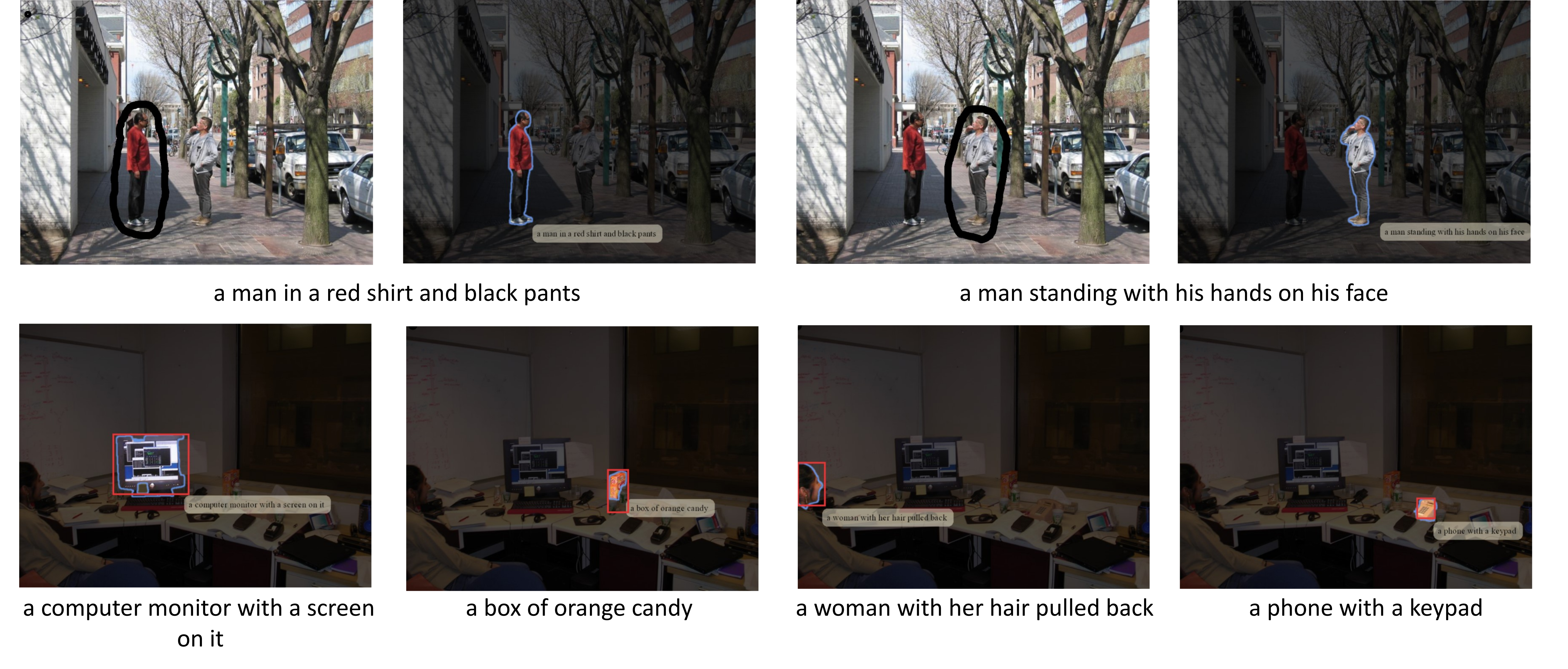}
    \caption{Visualization of describing the image with visual controls (trajectory and box).}
    \label{fig:example2}
\end{figure*}

\paragraph{Visual Controls.} 
As shown in Fig.~\ref{fig:example1}, we present some qualitative results that showcase the remarkable visual control capabilities of CAT. By placing click-point prompts in various locations within the images, CAT is able to accurately identify and describe corresponding objects, demonstrating its exceptional ability to caption a diverse range of objects in any given image. Furthermore, as shown in Fig.~\ref{fig:example2}, CAT's visual controls can be trajectory-based or bounding box-based, further highlighting its versatility and adaptability in generating accurate descriptions for a wide range of image content.

\begin{figure*}
    \centering
    \includegraphics[width=1.0\textwidth]{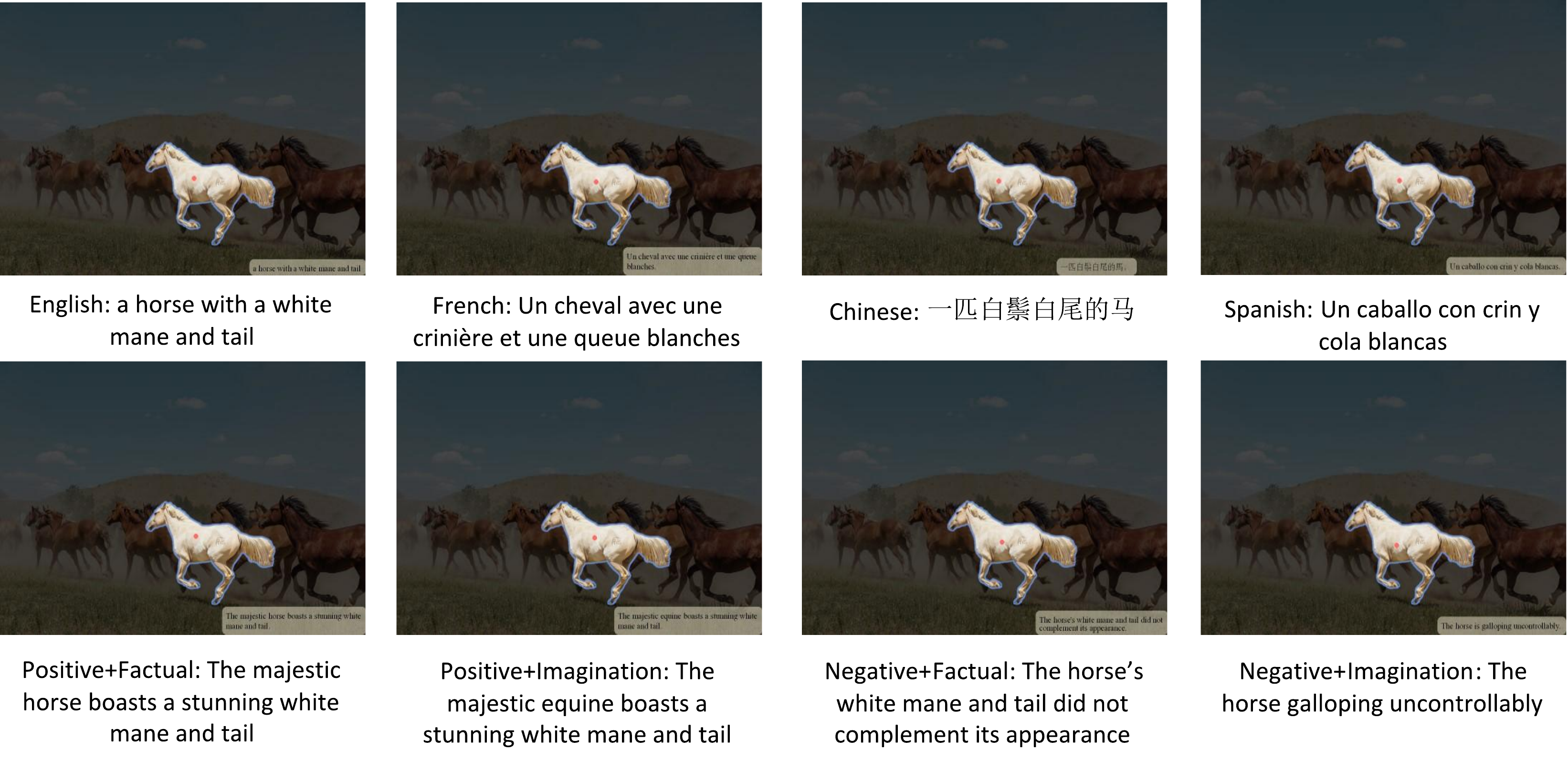}
    \caption{Visualization of generated captions with language controls (first row: multiple languages; second row: sentiment and factuality).}
    \label{fig:example3}
\end{figure*}

% \vspace{-1.0 em}
\paragraph{Language Controls.} 
We provide additional qualitative results that showcase the impressive language control capabilities of our model, as shown in Fig.~\ref{fig:example3}. Leveraging its advanced language controls (\ie, sentiment and factuality), CAT can generate captions with a diverse range of language styles, ranging from casual and conversational to formal and informative. These results highlight the model's remarkable ability to adapt its language output based on the specific needs and preferences of the user.


\begin{figure*}
    \centering
    \includegraphics[width=1.0\textwidth]{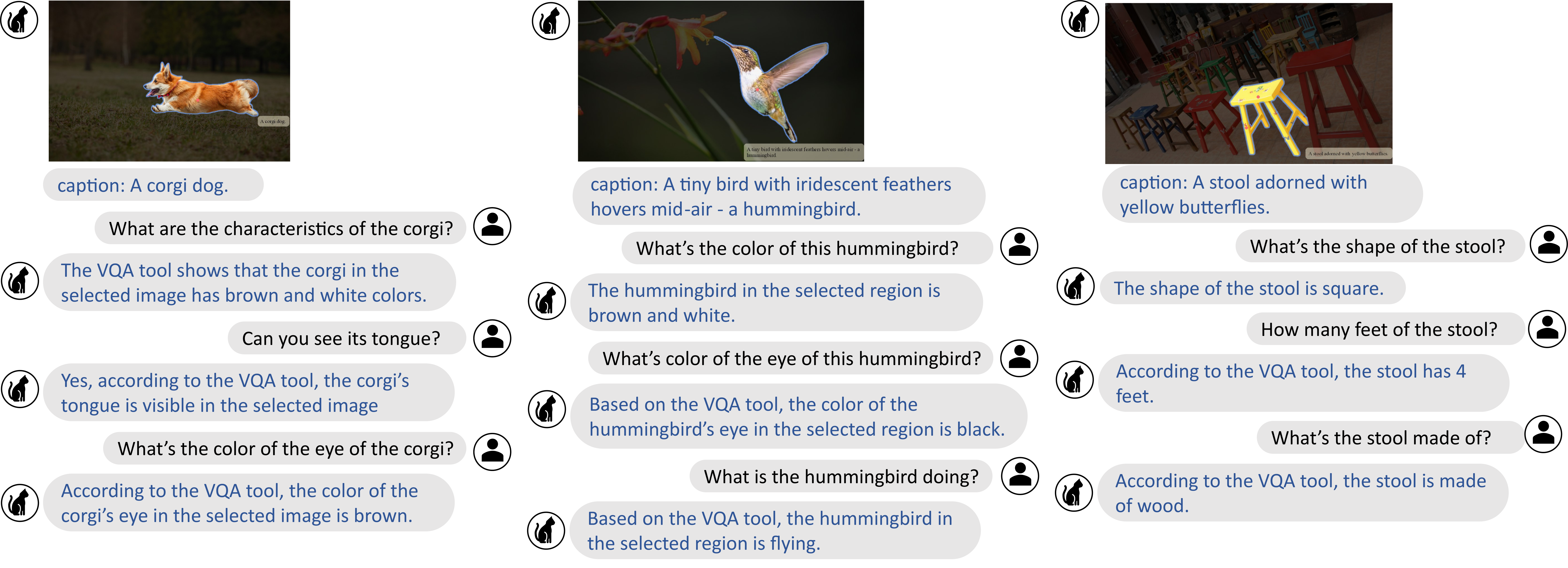}
    \caption{Examples of object-centric chatting.}
    \label{fig:example4}
\end{figure*}

\begin{figure*}
    \centering
    \includegraphics[width=1.0\textwidth]{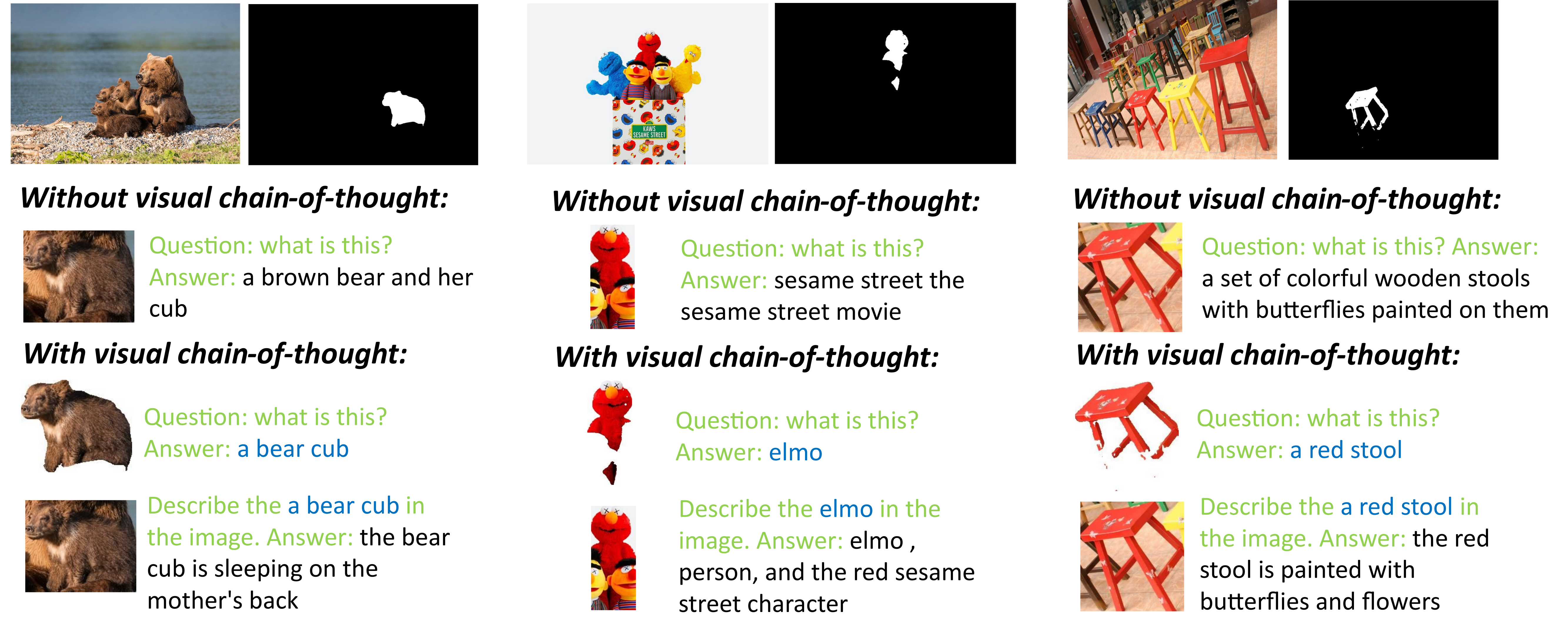}
    \caption{Examples of visual chain-of-thought.}
    \label{fig:example5}
\end{figure*}

\begin{figure*}
    \centering
    \includegraphics[width=1.0\textwidth]{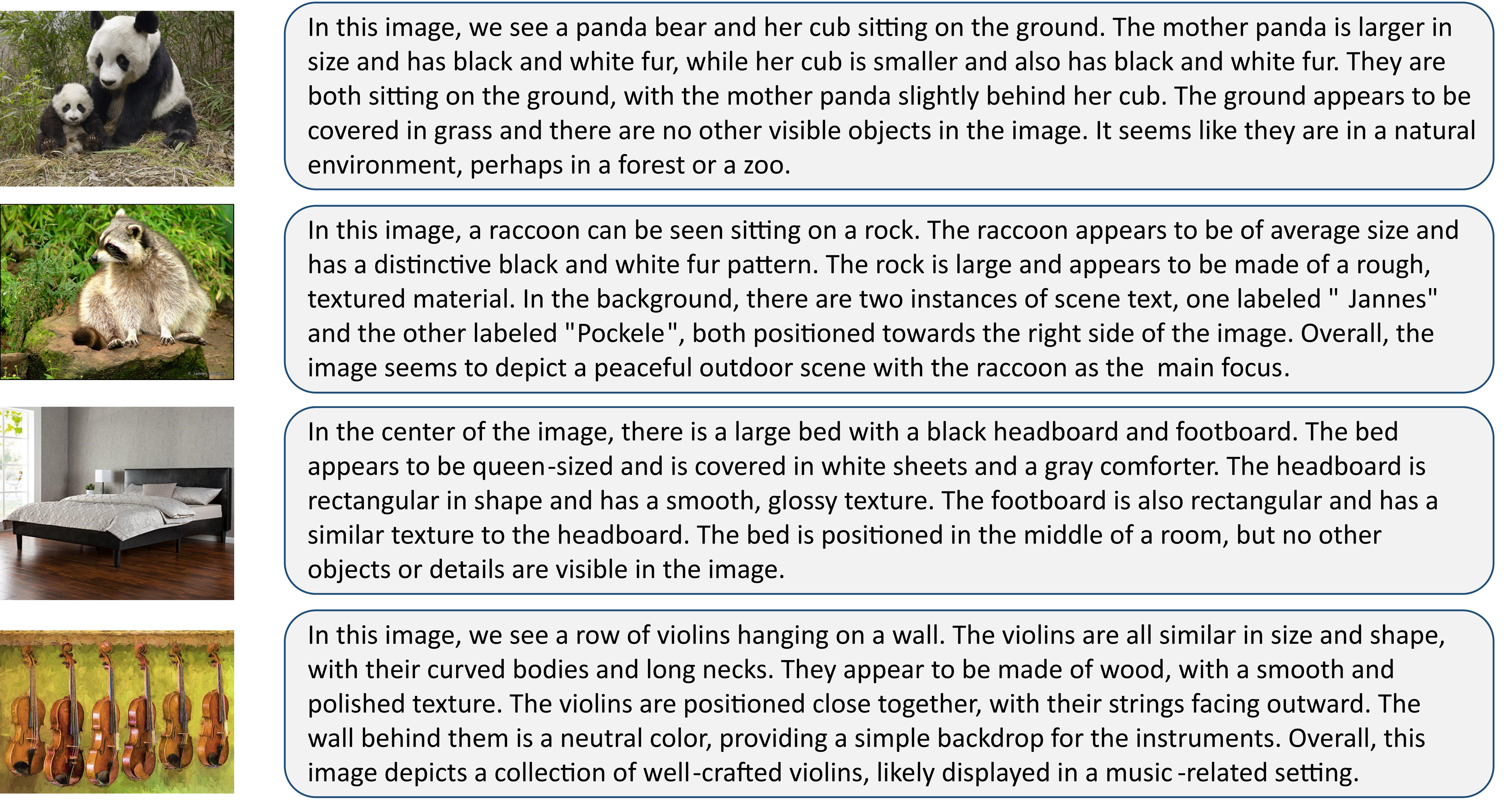}
    \caption{Examples of image paragraph captioning.}
    \label{fig:example6}
\end{figure*}

% \vspace{-1.0 em}
\paragraph{Object-Centric Chatting.} 
In Fig.~\ref{fig:example4}, we present further evidence of CAT's exceptional object-centric chatting ability. Specifically, our model is capable of performing visual question answering focused around selected objects within an image, showcasing its remarkable ability to engage in rich and meaningful conversations centered around specific visual content. 

% \vspace{-1.0 em}
\paragraph{Visual Chain-of-Thought.} Fig.~\ref{fig:example5} presents a variety of examples illustrating the efficacy of the visual CoT. As shown in this figure, performing direct inference on images containing backgrounds tends to be adversely influenced by the background content, thus hindering the captioner's ability to concentrate on the interested object. By incorporating a step-by-step thought process, the generated captions exhibit enhanced focus. Concurrently, the visual chain-of-thought facilitates the disclosure of more intricate details pertaining to the object of interest, thereby enabling the entire system to more effectively align with the user's intent.
% \vspace{-1.0 em}

\paragraph{Caption Everything in a Paragraph.}
Fig.~\ref{fig:example6} presents several examples of the resultant paragraphs. These generated paragraphs encompass the majority of the objects and textual information in the scene and even include some reasoning information. The overall descriptions provided by the paragraphs are accurate in general and logically coherent.
\section{Conclusion}
In this paper, we propose a foundation model augmented controllable image captioning framework, Caption AnyThing (CAT), that addresses the limitations of existing CIC approaches. Our proposed framework leverages pre-trained image captioners and integrate them with the SAM and an instruction-tuned LLM, thereby mitigating the reliance on human-annotated data and expanding the range of supported control signals. The unified representation of control signals in CAT enhances the model's flexibility and scalability, making it easily adaptable for interactive use and extension with new aspects of controllability. The experiments demonstrate that CAT provides a training-free and adaptable solution for controllable image captioning tasks, offering strong user-interactive capabilities.
\bibliographystyle{plain}
\bibliography{egbib}
\end{document}